\documentclass{article}
\usepackage{spconf}

\usepackage{cite}
\usepackage{amscd,amsmath,amssymb,amsfonts,latexsym,mathrsfs,amsthm,mathtools}
\usepackage{bm}
\usepackage{algorithm}
\usepackage{algpseudocode}
\usepackage{graphicx}
\usepackage{textcomp}
\usepackage{xcolor}
\usepackage{tikz-cd}
\usepackage{subcaption}
\usepackage{hyperref}
\usepackage[capitalize]{cleveref}
\usepackage{balance}
\usepackage{float}
\usepackage{subcaption}
\usepackage{siunitx}

\ninept

\newcommand{\V}[1]{\boldsymbol{\mathbf{#1}}}

\def\vtheta{{\bm{\theta}}}
\newcommand{\veta}{\bm{\eta}}

\newcommand{\y}{{\bf y}}
\newcommand{\x}{{\bf x}}
\newcommand{\X}{{\bf X}}
\newcommand{\vphi}{\bm{\phi}}

\newcommand{\R}{\mathbb{R}}
\newcommand{\D}{\mathcal{D}}

\newcommand{\N}{\mathcal{N}}

\def\vphi{{\bm{\phi}}}
\def\vPhi{{\bm{\Phi}}}

\begin{document}

\title{
% Robust, Scalable, and Online Gaussian Processes for Dynamic and Decentralized Settings
Robust, Online, and Adaptive Decentralized Gaussian Processes}

\name{Fernando Llorente${^1}$, Daniel Waxman${^2}$, Sanket Jantre${^1}$, Nathan M. Urban${^1}$, Susan E. Minkoff${^1}$}
\address{${^1}$ Applied Mathematics Department,
%Computing \& Data Sciences Directorate, 
Brookhaven National Laboratory
%, Upton, NY
\\${^2}$ Department of Electrical and Computer Engineering, Stony Brook University
%, Stony Brook, NY
}

% \author{Fernando Llorente$^{\star}$ \qquad Daniel Waxman$^{\dagger}$ \qquad Sanket Jantre$^{\star}$ \qquad Nathan M. Urban$^{\star}$\qquad  Susan E. Minkoff$^{\star}$ \\[0.6em]
% $^{\star}$Applied Mathematics Department, Computing \& Data Sciences Directorate, Brookhaven National Laboratory, Upton, NY \\[0.2em]
% $^{\dagger}$Department of Electrical \& Computer Engineering, Stony Brook University, Stony Brook, NY}

% \thanks{
% {
% % $^*$ Equal contribution.
% % \newline
% This work was supported by the National Science Foundation (NSF) under Award Number 2212506.
% }}
% }

% \author{
%     \IEEEauthorblockN{
%         Fernando Llorente$^\dagger$, 
%         Daniel Waxman$^*$, 
%         Sanket Jantre$^\dagger$, 
%         Sue Minkoff$^\dagger$
%     }
%     \\
%     \IEEEauthorblockA{
%         \begin{tabular}{cc}
%             \begin{tabular}{c}
%                 $^\dagger$Applied Math Department, CDS \\
%                 \textit{Brookhaven National Laboratory}
%             \end{tabular}
%             &
%             \begin{tabular}{c}
%                 $^*$Department of Electrical and Computer Engineering \\
%                 \textit{Stony Brook University}
%             \end{tabular}
%         \end{tabular}
%     }
% }

\maketitle

\begin{abstract}
Gaussian processes (GPs) offer a flexible, uncertainty‐aware framework for modeling complex signals, but scale cubically with data, assume static targets, and are brittle to outliers, limiting their applicability in large-scale problems with dynamic and noisy environments. Recent work introduced decentralized random Fourier feature Gaussian processes (DRFGP), an online and distributed algorithm that casts GPs in an information-filter form, enabling exact sequential inference and fully distributed computation without reliance on a fusion center. In this paper, we extend DRFGP along two key directions: first, by introducing a robust-filtering update that downweights the impact of atypical observations; and second, by incorporating a dynamic adaptation mechanism that adapts to time-varying functions. The resulting algorithm retains the recursive information-filter structure while enhancing stability and accuracy. We demonstrate its effectiveness on a large-scale Earth system application, underscoring its potential for in-situ modeling.
\end{abstract}

\begin{keywords}
Gaussian processes, 
random features, 
robust learning, decentralized inference, Earth system modeling.
\end{keywords}

% {\color{blue}[DW: Regarding the title: this paper isn't really federated or non-stationary... Federated learning is typically centralized, whereas our problem is significantly more difficult (I don't have a problem with \emph{decentralized federated settings}, but this becomes quite verbose, without much more information than just ``decentralized.'' And we don't really address non-stationarity much in the current batch of experiments.]}

\section{Introduction}

\begin{figure}[t]
    \centering
    \includegraphics[width=0.9\linewidth]{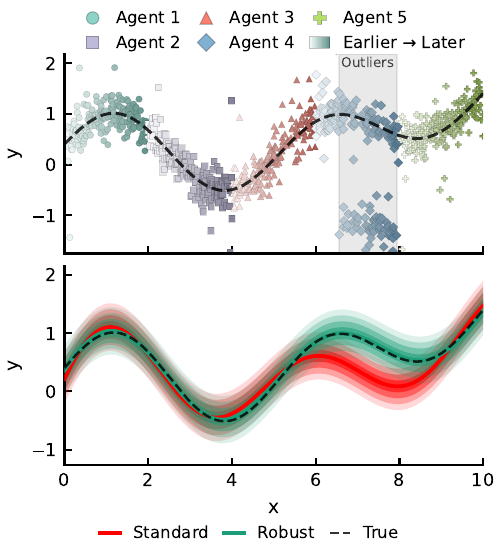}
    \caption{An illustration of our proposed online, decentralized, and robust GP inference approach. \textbf{Top:} Five agents receive data sequentially (with a data point's shade representing its temporal proximity). Agent 4's data is contaminated by localized random outliers. \textbf{Bottom:} The standard online decentralized RF-GP \cite{llorente24dynamic} becomes inaccurate due to these outliers, but our robust method, using tools from robust filtering theory, maintains superior performance.}
    \label{fig:marquee_figure}
\end{figure}

Gaussian processes (GPs) are widely used in signal processing and machine learning due to their ability to capture complex dependencies with principled uncertainty quantification \cite{rasmussen2006gaussian}. However, standard GP inference incurs cubic cost in the number of observations, assumes static functions, and is sensitive  to outliers. These limitations are further amplified in modern applications where data arrive sequentially, are distributed across several sensors, and both computation and communication resources are limited \cite{djuric2018cooperative}. In such settings, the model must learn online, adapt to time-varying  environments, and facilitate  robust inference.

In recent years, researchers have adapted GPs to use online approximations to enable sequential updates \cite{bui2017streaming, maddox2021conditioning}, distributed algorithms based on product-of-experts fusion to scale to large datasets \cite{deisenroth2015distributed, liu2018generalized}, and decentralized approaches to training and inference in multi-agent systems \cite{kontoudis2021decentralized, kontoudis2024scalable, hoang2019collective}. Robust filtering \cite{waxman24robust, duran2024outlier, chang2017unified, laplante2025robust} and dynamic kernels \cite{nyikosa2018bayesian, van2012kernel, van2012estimation, llorente24dynamic} further extend GPs to handle outliers  and temporal evolution. Online kernel machines such as random--feature methods and distributed consensus updates--address similar goals of scalable, streaming learning but lack principled, calibrated uncertainty quantification \cite{shen2019random, hong2021distributed, hong2021communication}. At the application level, decentralized and distributed GPs have been used for cooperative control of multi-agent systems \cite{lederer2022cooperative}, distributed environmental mapping and sensing \cite{di2022distributed}, collective online learning in large-scale systems \cite{hoang2019collective}, Earth system modeling \cite{rumsey2022hierarchical, grosskopf2021insitu}, and spatiotemporal traffic modeling and mobility-on-demand systems \cite{chen2015gaussian}.

Despite this progress, existing decentralized GP algorithms fail to meet at least one of the following desiderata: (a) approximation is clearly controlled via hyperparameters; (b) online inference is possible; (c) they are easily adaptable to time-varying or spatiotemporal environments; and (d) they are robust to contamination via outliers.

A step toward these goals is the {\it decentralized random Fourier feature Gaussian process} (DRFGP) \cite{drfgp25} which uses random Fourier features to map inputs into a finite-dimensional embedding, approximating a GP with a stationary kernel by a Bayesian linear model \cite{rahimi2007random, lazaro2010sparse}. Exact and recursive updates are then available via Kalman filtering. Using natural parameters, additive consensus provides a decentralized solution, where agents learn single, shared GP model, in contrast to product-of-experts schemes that do not preserve a coherent global GP \cite{deisenroth2015distributed, liu2018generalized}. We further discuss the DRFGP in \cref{sect_drfgp}. The DRFGP can be viewed both as a decentralization of online GPs via basis expansions, \cite{waxman2024doebe}, or as a kernelized analogue of distributed Bayesian linear regression that aggregates sufficient statistics by consensus \cite{wang2015distributed}.

In this paper, we extend DRFGP with two simple yet effective enhancements: (i) per-agent, per-sample robust updates that downweight atypical observations, inspired by the robust filtering literature; and (ii) an explicit dynamic model that adapts to time-varying functions. Together, these yield the \emph{robust, online, and adaptive decentralized GP (ROAD-GP)}. The enhancements are straightforward to integrate and, in practice,  substantially improve stability and accuracy. We demonstrate the approach in an Earth system application that assimilates large volumes of streaming data, showing that ROAD-GP enables effective sequential, distributed inference.

\section{Decentralized RF-based Gaussian Processes}
\label{sect_drfgp}

We first review the DRFGP \cite{drfgp25}, recalling how random Fourier feature GPs admit a Bayesian linear model form with an analytical posterior. We then present the information-filter formulation, which naturally enables online and distributed inference and underpins our robust and dynamic extensions.

\subsection{Fusion Center Solution}

We begin with the centralized setting, in which all data are available at a single location. This setting serves as the reference solution that decentralized algorithms aim to approximate.
We want to learn a function $f(\x)\in\R,\ \x\in\R^d$ using a GP with a stationary kernel $k(\V{x}, \V{x}')$. In the random feature GP (RF-GP), the stationary kernel of a GP is approximated using random Fourier features \cite{rahimi2007random}, i.e., $k(\x,\x') \approx \vphi(\x)^\top \vphi(\x')$ where $\vphi(\x)\in\R^{2J}$ is defined as
{\footnotesize
\begin{align*}
    \vphi(\x) = \frac{1}{\sqrt{J}} \left[\sin(\x^\top \V v_1), \, \cos(\x^\top \V v_1),\,...\, ,\sin(\x^\top \V v_J)\, ,\cos(\x^\top \mathbf{v}_J)\right]^\top
\end{align*}}

This mapping is computed using samples from the spectral density of $k(\x,\x')$ \cite{rahimi2007random,lazaro2010sparse}.
Then the RF-GP approximation of $f(\x)$ reduces to a Bayesian linear model with parameters $\vtheta \in \R^{2J}$,
% \begin{align}\label{eq_RF_GP}
  $ f(\x) = \vphi(\x)^\top\vtheta, $
% \end{align}
with prior $\vtheta \sim \N({\bf 0},\sigma_{\theta}^2{\bf I}_{2J})$. 
Given a dataset of observations of $f(\x)$ corrupted by Gaussian noise with variance $\sigma^2_\text{obs}$, such that $\D = (\X, \y)$ with $\X\in\R^{T\times d}$ and $\y\in\R^T$. 
The posterior of $f(\x)$ is determined by the posterior of $\vtheta$, which is also Gaussian. Denoting $\vPhi = [\vphi(\x_1) \dots \vphi(\x_T)] \in \R^{2J\times T}$, the posterior of $\vtheta$ is $p(\vtheta \mid \X, \y) = \mathcal{N}(\vtheta \mid \bm{\mu}_c, \bm{\Sigma}_c)$, with
\begin{equation} \label{eq:rff_gp_posterior_quantities}
\bm{\mu}_c = \frac{1}{\sigma_\text{obs}^2}{\bf \Sigma}_c \vPhi \y, \qquad {\bf \Sigma}_c = \left( \frac{1}{\sigma_\text{obs}^2}\vPhi \vPhi^\top + \frac{1}{\sigma_\theta^2} {\bf I} \right)^{-1}.
\end{equation}
It will be convenient later to express the above in an equivalent information form. To this end, let ${\bf D}_c = {\bf \Sigma}_c^{-1}$ and $\veta_c = {\sigma_\text{obs}^{-2}}\vPhi\y$, then \cref{eq:rff_gp_posterior_quantities} can be rewritten as
\begin{equation} \label{eq:posterior_w_information_quantities}
    \bm{\mu}_c = {\bf D}_c^{-1} \veta_c, \qquad {\bf D}_c = \frac{1}{\sigma_\text{obs}^2}\vPhi \vPhi^\top + \frac{1}{\sigma_\theta^2} {\bf I}.
\end{equation}
This approximation provides the baseline centralized solution, with complexity $\mathcal{O}(TJ^2+J^3)$, compared with the $\mathcal{O}(T^3)$ scaling of vanilla GPs \cite{rasmussen2006gaussian}, thereby reducing computational cost when $T \gg J$.

% \subsection{Information Filter}

While the closed-form solution is exact, it assumes the data are centralized. To make this approach both online and distributed, we exploit its additive structure. The {\it information} quantities $\veta_c$ and ${\bf D}_c$ admit an additive form in the case of conditionally independent Gaussian observations: 
\begin{align}
    {\bf D}_c &= \sum_{t=1}^{T} {\bf P}_t = 
    \frac{1}{\sigma_\theta^2}{\bf I} + 
    \sum_{t=1}^{T} 
    \left(\frac{1}{\sigma_\text{obs}^2} \vPhi_t \vPhi_t^\top
    \right), \label{eq:D_c_sum}
    \\
    \veta_c &= \sum_{t=1}^{T} {\bf s}_t = \sum_{t=1}^{T} \frac{1}{\sigma_\text{obs}^2} \vPhi_t \y_t\ , \label{eq:eta_c_sum}
\end{align}
where $\vPhi_t\in\mathbb{R}^{2J\times N_t}$, $\y_t\in\mathbb{R}^{N_t}$ and $N_t$ denotes the number of observations in the $t$-th batch. For $N_t=1$, we have $\vPhi_t = \vphi(\x_t)\in\R^{2J}$ and $\y_t = y_t\in \R$. This decomposition opens the path for online and distributed learning. In the online setting, new data arrive at each time instant $t$, updating the quantities as follows:
\begin{align}\label{eq_fusion_center_online1}
    {\bf D}_{c,t} &= {\bf D}_{c,t-1} + {\bf P}_{t} 
    \\
    \veta_{c,t} &= 
     \veta_{c,t-1} + {\bf s}_{t},
\label{eq_fusion_center_online2}
\end{align}
where we define the incremental quantities ${\bf P}_t$ and ${\bf s}_t$ as
\begin{align}
    {\bf P}_t = \frac{1}{\sigma_\text{obs}^2} \vPhi_t \vPhi_t^\top, &\quad {\bf P}_0 = \frac{1}{\sigma_{\theta}^2}{\bf I}. \\
    {\bf s}_t = \frac{1}{\sigma_\text{obs}^2} \vPhi_t \y_t, &\quad {\bf s}_0 = {\bf 0}.
\end{align}
In summary, rewriting the posterior in information form reveals its recursive structure. Specifically, every new batch of observations contributes an additive update to  $\veta_t$ and ${\bf D}_t$, as per Eqs. \eqref{eq_fusion_center_online1}-\eqref{eq_fusion_center_online2}, enabling online learning without storing all past data. 

\subsection{Decentralized multi-agent systems}
\begin{figure}[t]
    \centering
    \includegraphics[width=0.55\linewidth]{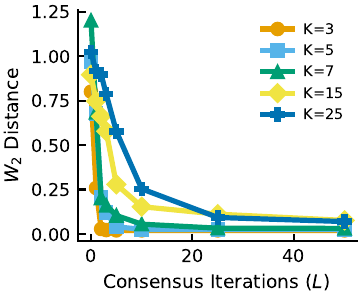}
    \caption{An illustration of consensus building: $K \in \{3, 5, 7, 15, 25\}$ agents are connected in a ``ring'' topology and perform $L$ consensus iterations. The figure shows the 2-Wasserstein distance to the centralized posterior, which decreases rapidly as $L$ increases.}
    \label{fig:consensus}
\end{figure}
The additive formulation also applies when data are spread across N agents at each time instant $t$. Introducing a subscript $k$ to denote the $k$-th agent, the update equations become
\begin{align}
% \label{eq_fusion_center_online}
    {\bf D}_{c,t} &= 
    {\bf D}_{c,t-1} +  {\bf P}_{t}
    =
    {\bf D}_{c,t-1} + \sum_{k=1}^{K} {\bf P}_{k,t}
    ,
    \\
    \veta_{c,t} &= 
     \veta_{c,t-1} + {\bf s}_{t}
     =
     \veta_{c,t-1} + \sum_{k=1}^K {\bf s}_{k,t}.
\end{align}
This update scheme corresponds to a distributed information filter, in which each agent at time $t$ computes the local quantities ${\bf P}_{k,t}$ and ${\bf s}_{k,t}$. The exact solution can be recovered by aggregating the agents' quantities to obtain ${\bf P}_{t} = \sum_{k=1}^K{\bf P}_{k,t}$ and ${\bf s}_{t} = \sum_{k=1}^K{\bf s}_{k,t}$. 
However, in a fully distributed scenario, no fusion center exists, and agents can only communicate with their neighbors.

If we want a \emph{fully} decentralized implementation, the goal is thus to approximate the sums ${\bf P}_{t} = \sum_{k=1}^K{\bf P}_{k,t}$ and ${\bf s}_{t} = \sum_{k=1}^K{\bf s}_{k,t}$ that a fusion center would compute, while sharing information only with neighbors. Under mild conditions \cite{olfati2007consensus}, this goal can be achieved via additive consensus algorithms, in which each agent approximates $\widetilde{{\bf P}}_{k,t}^{(L)}\approx {\bf P}_{t}$ and $\widetilde{{\bf s}}^{(L)}_{k,t} \approx {\bf s}_{t}$ for all $k$. Here, a tilde sign denotes the consensus approximation at agent $k$ after $L$ communication rounds. Each agent then acts approximately as a fusion center that computes the full posterior up to time $t$. The convergence of GP predictions to the centralized posterior under additive consensus is shown in \cref{fig:consensus}.

\subsection{Ensembles}
Thus far, we have described decentralized inference for a single RF-GP. In practice, performance depends on kernel hyperparameters such as length scales. To handle hyperparameter tuning in DRFGP, we use an ensemble of $M$ RF-GPs at each agent (e.g. by sampling length scales from a prior). The agent network then aims to approximate the mixture of RF-GPs at a fusion center. In this case, even with a fusion center, we do not recover the exact Bayesian model averaging (BMA) weights unless the data batches are independent--a common assumption in distributed GPs \cite{liu2018generalized}. Nevertheless, ensembles provide a simple way to further improve robustness and predictive performance of the final model.

\section{Novel schemes}
Having introduced DRFGP, we now present two extensions that make it suitable for dynamic environments and robust to outliers. These extensions are easy to incorporate and improve stability and accuracy in challenging scenarios.

\subsection{Time-varying DRFGP}
The DRFGP introduced in Section \ref{sect_drfgp} assumes a static regression function $f(\x)$, and hence does not handle time-varying functions. In the centralized setting, one solution is to impose a state transition on the model parameters $\vtheta$ (e.g., a random walk), a popular way to introduce temporal drift. This dynamical process turns the original Bayesian model into a state-space model (SSM),
\begin{align}\label{eq_rw_evolution}
    \vtheta_{t} &= \vtheta_{t-1} + {\bf u}_{t} \\
    y_t &= \vphi(\x_t)^\top \vtheta_t + {\bf n}_t,\quad t=1,2,\dots,
\end{align}
where ${\bf u}_t$ and ${\bf n}_t$ are independent Gaussian random variables for all $t$. The back-to-prior (B2P) and uncertainty-injection (UI) can be derived from Eq. \eqref{eq_rw_evolution} \cite{van2012estimation,llorente24dynamic}. In practice, B2P and UI forgetting corresponds to downweighting the (per-agent) quantities ${\bf D}_{t-1}$ and $\veta_{t-1}$ using a forgetting coefficient $\nu\in[0,1]$ before updating with the new batch of observations.

Alternatively, we can keep the Bayesian model static, and enlarge our input vector $\x$ with a variable $t$ so that we learn the dynamic function directly, $f(\x,t) = \phi(\x,t)^\top \vtheta$. Let $\widetilde{\x} = [\x,t]\in\R^{d+1}$, where $\x$ denotes the {\it spatial}\footnote{$\x$ is not required to be a spatial variable in the strict physical sense; $\x$ are the parameters of the function we aim to learn, but we assume the function itself changes with time.} variable and $t$ denotes time. The mapping $\vphi(\widetilde{\x})$ is equivalent to using a stationary {\it spatio-temporal} kernel,
\begin{align}
   k_{\text{st}}(\widetilde{\x},\widetilde{\x}') = k_{\text{s}}(\x,\x')\times k_{\text{t}}(t,t') \approx \vphi(\widetilde{\x})^\top \vphi(\widetilde{\x}').
\end{align}
There is a natural connection between the forgetting mechanism and the use of temporal kernels. For instance, B2P forgetting corresponds to using an Ornstein-Uhlenbeck temporal kernel \cite{van2012estimation}. In practice, we find the spatio-temporal kernel approach simpler and more effective, allowing us to proceed exactly as in the static case while implicitly capturing temporal correlations.

\subsection{Robust DRFGP}
Another practical challenge is robustness. The RF-GP with Gaussian likelihood can be highly sensitive to anomalies. This is particularly important in decentralized systems because a single agent receiving biased observations can contaminate the entire network. To improve robustness, we adapt tools from robust filtering theory and M-estimation, in particular the unified form of \cite{chang2017unified}, which is similar to the generalized Bayes updates of \cite{duran2024outlier}. Similar robust filters have recently demonstrated success in the centralized sequential GP literature \cite{waxman24robust,laplante2025robust}.

In a {\it robust} distributed information filter, each local agent computes a diagonal weight matrix  ${\bf W}_{k,t} = \text{diag}(w^{(i)}_{k,t})$. The weight $w^{(i)}_{k,t}$ is a function of the standardized residual  $ e^{(i)}_{n,t}=(y^{(i)}_{k,t} - \widehat{y}^{(i)}_{k,t})/\sigma_{y,k,t}$, where $\widehat{y}_{k,t}$ and   $\sigma_{y,k,t}$ denote the mean prediction and its standard deviation computed at agent $k$ for $i=1,\dots,K_t$. The update equations are now
\begin{align}
& {\bf D}_{t} \leftarrow {\bf D}_{t-1} + \sum_{k} \frac{\vPhi_{k,t} {\bf W}_{k,t} \vPhi_{k,t}^{\top}}{ \sigma_{\text{obs}}^2} \\
& \veta_t \leftarrow \veta_{t-1}+\sum_{k} \frac{\vPhi_{k,t}^{\top} {\bf W}_{k,t}^{\top} \y_{k,t}}{\sigma_{\text{obs}}^2}
\end{align}
Weights $w^{(i)}_{k,t}$ are derived from robust weighting functions. For instance, the Huber and Hampel strategies are defined as
\begin{align}
w^{(i)}_\text{Huber}(e) &= 
\begin{cases}
1, & |e|\leq \delta, \\
\delta/|e|, & |e|>\delta,
\end{cases} \\
w^{(i)}_\text{Hampel}(e) &= 
\begin{cases}
1, & |e|\leq a, \\
a/|e|, & a < |e| \leq b, \\
c-|e|\,/\, (c-b), & b<|e|\leq c, \\
0, & |e|>c,
\end{cases}
\end{align}
These updates are equivalent to carrying out inference under a tempered likelihood that downweights atypical observations.
% contribute less to the posterior.

% \begin{algorithm}[t]
% \caption{Decentralized RF-GPs at Agent $n$} \label{alg:coopbayesopt_known_hypers}
% \begin{algorithmic}[1]
% \State \textbf{Initialization:} $\Tilde{\mathbf{P}}^{(L)}_{n,0}$ and $\Tilde{\V{s}}^{(L)}_{n,0}$ %for $n=1,\dots, N$
% ; hyperparameters $\sigma_{\V\theta}^2, \ell, \sigma_{\text{obs}}^2$.
% \For{$t = 1 : T$}
%     \State \textbf{Step 1:} Acquire new data $(\x_{n, t},\y_{n,t})$ and compute ${\bf P}_{n, t}$, ${\bf s}_{n, t}$.
%     \State \textbf{Step 2:} Obtain $\Tilde{{\bf P}}^{(L)}_{n, t}$ and $\Tilde{{\bf s}}^{(L)}_{n, t}$ by peforming $L$ iterations of consensus.
%     % \cref{eq:P_consensus,eq:s_consensus}.
%     % Exchanging of $\mathbf{P}_{i,t}$ and $\mathbf{s}_{i,t}$ with neighbors and compute $\bar{P}^{(L)}_{n,t}$ and $\bar{s}^{(L)}_{n,t}$ via consensus in Eq. \eqref{eq_consensus_P_and_s:a}-\eqref{eq_consensus_P_and_s:b}.
%     \State \textbf{Step 3:} Update the states $\mathbf{D}_{n,t}$ and $\mathbf{\eta}_{n,t}$ based on consensus estimates according to \cref{eq:D_n_update,eq:eta_n_update}.
%     \State \textbf{Step 4:} Report the predictive estimate $\widehat{p}_n(\vtheta|\D_t) = \mathcal{N}\left({\bf D}^{-1}_{n,t}\veta_{n,t}, {\bf D}^{-1}_{n,t}\right)$.
% \EndFor
% \end{algorithmic}
% \end{algorithm}

\section{Experiments}
We now illustrate the proposed extensions in an Earth-system application.
The goal is not to solve a specific scientific problem but to empirically validate ROAD-GP in a realistic scenario where (i) there is large-scale streaming data, (ii) there may be data contamination and non-stationarity, and (iii) distributed computation and communication efficiency  are desirable. We focus on two main questions. First, do the robust modules prevent the network from being contaminated when an agent receives extreme observations? Second, does consensus allow the network to converge to a common approximation in decentralized settings?
% First, we show that robust strategies are important for yielding a stable approximation when the data are corrupted with extreme observations.
% Second, we investigate several variants of R-DRFGP, namely, considering different hyperparameters such as the number of consensus iterations, the number of models, and different dynamics.

\begin{figure*}[t]
    \centering
    \begin{subfigure}[t]{0.7\textwidth}
        \includegraphics[width=\linewidth]{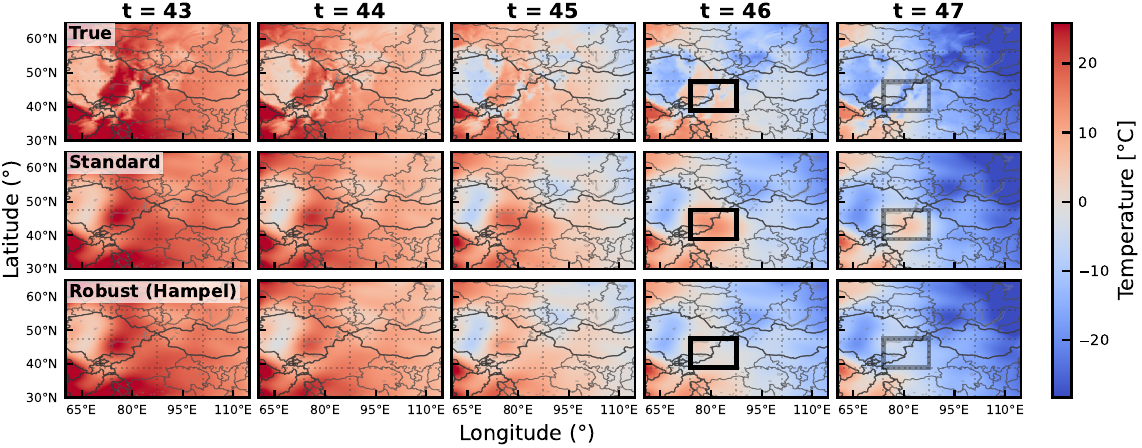}
        \caption{}\label{fig:weather_results_map}
    \end{subfigure}%
    \hspace{7.5mm}
    \begin{subfigure}[t]{0.24\textwidth}
        \includegraphics[width=\linewidth]{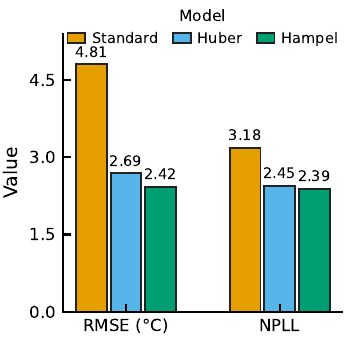}
        \caption{}\label{fig:robust_metrics}
    \end{subfigure}
    \caption{\textbf{(a)} Visualizations of spatiotemporal predictions in the ``robustness to outliers'' experiment. Grid lines denote agent boundaries, and squares indicate the region where outliers are injected at $t=46$. Note the effect on predictions at $t = 47$. \textbf{(b)} RMSE and negative predictive log-likelihood (NPLL) computed over the final time instant ($t=47$).}
    \label{fig:robust_weather_experiment}
\end{figure*}

\begin{figure}
    \centering            
    \includegraphics[width=0.95\linewidth]{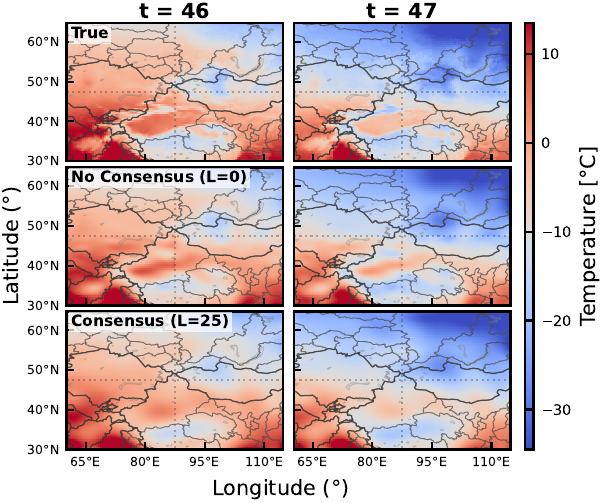}
    \caption{Predictions of the spatiotemporal field with and without the use of additive consensus. In ``No Consensus,'' each agent's local GP is stitched together; \cref{fig:consensus_predictions_global} visualizes each agent's global predictions.}
    \label{fig:consensus_predictions_ts}
\end{figure}

\begin{figure}
    \centering
    \includegraphics[width=0.95\linewidth]{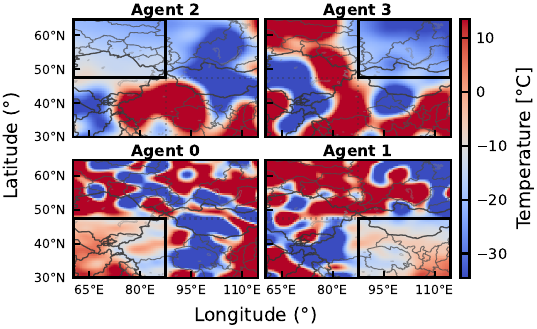}
    \caption{Global estimates of each local agent in \cref{fig:consensus_predictions_ts}. Black rectangles denote the area in which each agent receives data.}
    \label{fig:consensus_predictions_global}
\end{figure}

\subsection{Experimental setup}

We use gridded weather data for a region of central Asia from the CRU dataset \cite{harris2020version}, processing monthly temperatures in streaming fashion.
%We simulate a sequential estimation scenario by processing monthly temperature datasets of a region in Central Asia in a streaming fashion, from  \hyperlink{CRU Dataset}{https://crudata.uea.ac.uk/cru/data/hrg/}.
We aim to highlight ROAD-GP's ability to ingest large volumes of streaming data and summarize them with fixed-size statistics, while yielding an approximation of the centralized solution with low communication overhead. We include ensembles with spatial lengthscales $\ell_s \in \{0.01, 0.05, 0.1\}$, a temporal lengthscale $\ell_t = 4.0$, observation variance $\sigma_\text{obs}^2 = 0.05$, and process variance $\sigma_\theta^2 \in \{ 1.0, 25.0 \}$. We normalize the spatial and output dimensions, but not the temporal dimension.
% {\color{red}$N_\text{agent} = 16$, $N_\text{batch}=459$, and $T=48$ }

\subsection{Robustness to outliers}
We first assess the robustness of ROAD-GP. We consider $K=4$ agents, with $J=\num{1000}$ random frequencies, each receiving $N_\text{batch}=459$ spatial observations over $T = 48$ successive months, for a total of $\approx \num{325000}$ data points. Inspired by \cite{laplante2025robust}, we inject extreme outliers at time $t=46$, biasing $30\%$ of the observations to random values $8$ standard deviations warmer than recorded -- similar to extremely anomalous events or sensor failure. We illustrate predictions for $t=43,\cdots,47$ for the standard DRFGP with a spatiotemporal kernel and ROAD-GP with Hampel weights in \cref{fig:weather_results_map}. ROAD-GP safely ignores the influence of these outliers, which is evident both visually and in the reported metrics of \cref{fig:robust_metrics}.

\subsection{Consensus and approximation to the centralized solution}
We next evaluate the effect of consensus, using four agents on the same dataset for better visualization. With $L=0$ communication rounds, the model reduces to a collection of independent (local) RF-GPs, one per agent. With sufficiently large $L>0$, each agent's posterior then converges to the global, centralized solution. \cref{fig:consensus_predictions_global} shows that local solutions ($L=0$) can produce crisper predictions, but at the cost of potentially poor global estimates.

\section{Conclusions}
We have extended the decentralized random Fourier feature Gaussian process (DRFGP) framework with two lightweight yet powerful modules: robust updates that mitigate the effect of outliers, and dynamic mechanisms that adapt to time-varying functions. These modifications preserve the recursive and decentralized structure of DRFGP, allowing agents to maintain constant-size statistics and collectively approximate the centralized solution. Using curated weather datasets, we have validated our approach and shown that the new modules can be vital for obtaining stable solutions under corrupted data streams. These results highlight the potential for robust, dynamic in-situ modeling of large-scale streaming data.

% \clearpage
\balance

\bibliographystyle{IEEEtran}
\bibliography{refs}

\end{document}